\documentclass[10pt,twocolumn,letterpaper]{article}

\usepackage[pagenumbers]{cvpr} % To force page numbers, e.g. for an arXiv version

\definecolor{cvprblue}{rgb}{0.21,0.49,0.74}
\usepackage[pagebackref,breaklinks,colorlinks,allcolors=cvprblue]{hyperref}

\usepackage{multirow}

\def\paperID{*****} % *** Enter the Paper ID here
\def\confName{CVPR}
\def\confYear{2025}

\title{Fine-Grained Open-Vocabulary Object Recognition via User-Guided Segmentation}

\author{Jinwoo Ahn \\
UC Berkeley \\
\and
Hyeokjoon Kwon \\
KAIST \\
\and
Hwiyeon Yoo \thanks{Corresponding Author} \\
Seoul National University \\
}

\begin{document}
\maketitle
\begin{abstract}
    Recent advent of vision-based foundation models has enabled efficient and high-quality object detection at ease. Despite the success of previous studies, object detection models face limitations on capturing small components from holistic objects and taking user intention into account. To address these challenges, we propose a novel foundation model-based detection method called \textbf{FOCUS}: \textbf{F}ine-grained \textbf{O}pen-Vocabulary Object Re\textbf{C}ognition via \textbf{U}ser-Guided \textbf{S}egmentation. FOCUS merges the capabilities of vision foundation models to automate open-vocabulary object detection at flexible granularity and allow users to directly guide the detection process via natural language. It not only excels at identifying and locating granular constituent elements but also minimizes unnecessary user intervention yet grants them significant control. With FOCUS, users can make explainable requests to actively guide the detection process in the intended direction. Our results show that FOCUS effectively enhances the detection capabilities of baseline models and shows consistent performance across varying object types.
    % both quantitatively and qualitatively.
\end{abstract}    
\section{Introduction}
\label{sec:intro}

\begin{figure}[t]
  \centering
   \includegraphics[width=\linewidth]{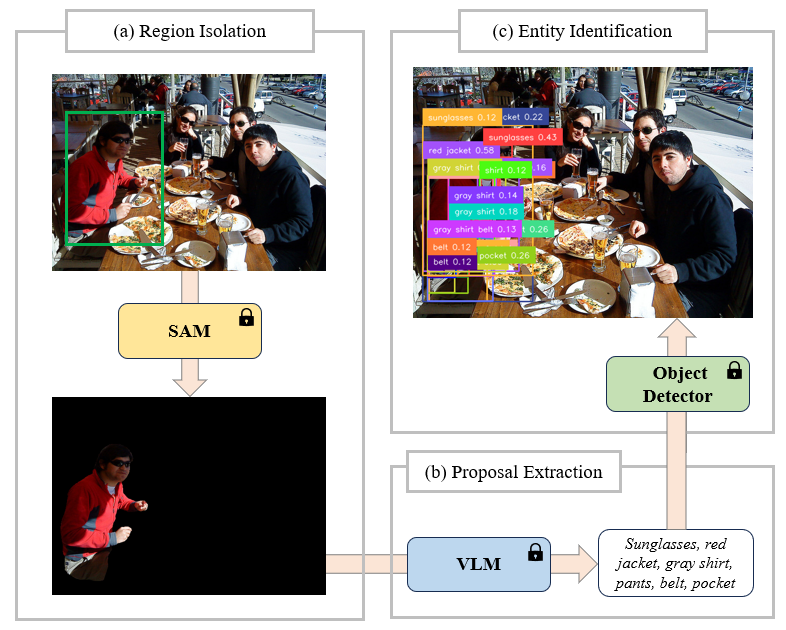}
   \caption{A \textbf{conceptual overview} of the FOCUS framework is shown above. (1) First, the input image and bounding box are processed to eliminate unnecessary regions from the image. (2) Second, the VLM is prompted to propose all objects that are present in the processed image. (3) Third, the proposal from the previous step serves as guidance to output final detection results.}
   \label{fig:pipeline_simple}
\end{figure}

% Open-vocabulary object detection aims to identify and localize objects from any category given an input image. More specifically, the term ``open-vocabulary" emphasizes the flexibility in recognizing objects beyond items that were part of its predefined labels, allowing consistent performance even when presented with previously unseen parts. Recent works in this domain distill the generalization capabilities of models that specialize in vision-language alignment \citep{liu2023grounding, li2021grounded, radford2021clip} to achieve promising performance.
% Enhancing the capabilities of open-vocabulary object detection models is essential, as accurately identifying instances from visual scenes forms the foundation of many core visual perception tasks such as autonomous navigation \citep{Pan_2024_CVPR, Wang_2024_CVPR} and vision-based robot learning \citep{blank2024scaling, murray2024teaching, shang2024theia}. Consider an agent designed to assist with tasks by responding to user requests. The user might ask the agent to prepare for an outdoor activity, like one illustrated in Figure \ref{fig:pipeline_simple}. To execute such a request, the agent would ideally need to identify and retrieve specific items like ``belts" and ``sunglasses" from a cluttered scene, rather than simply recognizing generalized labels like ``clothing" or ``accessories". This requires the agent to discern subtle, less prominent objects in close proximity, a level of detail essential for understanding and executing specific tasks tailored to user needs.

Aligning model outputs with human preferences is essential for enabling future systems to effectively carry out real-world tasks. This requires the models to interpret diverse, unconstrained user inputs, as limited language processing capabilities restrict users from fully expressing their intentions. Current object detection models demonstrate promising performance in accurately identifying a wide range of instances from visual scenes by distilling the generalization capabilities of models that specialize in vision-language alignment \citep{liu2023grounding, li2021grounded, radford2021clip}. In the future, however, these models may need to provide more detailed detection results based on more descriptive textual inputs that expand from simple object labels. Imagine a scenario where a user expects the model to detect highly granular components within a scene like the one from Figure \ref{fig:pipeline_simple}. The user would ideally expect the model to refrain from merely identifying broad categories like ``person" but expect it to pinpoint and retrieve specific items such as ``belts" and ``sunglasses" amidst the clutter. Detection models should, therefore, aim to possess the capability to understand user intent, enabling them to perform tasks such as discerning subtle, less prominent objects in close proximity. This direction would represent progress towards developing systems that move beyond basic detection, delivering meaningful insights applicable for humans in real-world scenarios.

Recognizing fine-grained components becomes increasingly challenging as visual scenes begin to contain more intricate details. These limitations arise because current models are designed to prioritize objects with the strongest features, which consequently leads to a limited understanding of compositionality, the idea that holistic objects always consist of meaningful constituent parts. Unfortunately, existing solutions developed to address these issues are inefficient and constrained, requiring manual intervention from the user. Even advanced detection models like GroundingDINO \citep{liu2023grounding} and YOLOWorld \citep{Cheng2024YOLOWorld} rely on users to specify the class labels of objects they wish to detect, restraining the models to function autonomously.

To address these limitations, we propose \textbf{FOCUS}: \textbf{F}ine-grained \textbf{O}pen-Vocabulary Object Re\textbf{C}ognition via \textbf{U}ser-Guided \textbf{S}egmentation. FOCUS offers outstanding performance on open-vocabulary object recognition tasks without relying on humans to label which objects to detect. It also allows users to actively guide the detection process in natural language by including vision language models (VLMs) in its pipeline, ensuring alignment between the detection results and the user intention. Alongside, the proposed approach allows users to incorporate alternative models in replacement of those in this paper across all of its modules. % We explain the structure of the pipeline and the details about the individual modules in Section \ref{sec:method}. 

More specifically, FOCUS is mainly comprised of three stages. It initially processes the image and the bounding box inputted from the user through the Segment Anything Model (SAM) \citep{kirillov2023segany, ravi2024sam2} to obscure regions that are not of interest to the user, as shown at the end of the Attentive Segmentation stage in Figure \ref{fig:pipeline_simple}. Then, the VLM takes the processed image to suggest objects for detection based on the user-specified natural language prompt. This prompt guides the model to propose only objects that align with the user request. For example, users could prompt ``detect all items that are smaller than an average human hand" if needed. This user-driven guidance allows flexible control over the scope and detail of object proposals. Lastly, existing detection models utilize the proposal from the previous stage to output the final detection results.

By integrating these foundation models into a pipeline, as outlined in Figure \ref{fig:pipeline_simple}, FOCUS effectively addresses the stated limitations, with each model contributing a specific function to streamline the downstream process. The use of SAM assists the detection models to \textit{focus} exclusively on the region of interest selected by the user, enabling fine-grained object detection even in complex scenes. Similarly, incorporating VLMs to the pipeline enables the users to make explainable requests in the detection process that is not limited in length or scope while automating the process of curating the list of objects to be detected.

Our main contributions can be summarized as follows: 
\begin{itemize}
    \item We present FOCUS, a zero-shot approach that leverages foundation models to more confidently identify granular components from complex scenes.
    \item The proposed method enables users to guide the detection process via natural language, supporting a variety of tasks and aligning the results with nuanced user intentions.
    \item The proposed method demonstrates consistent performance across diverse targets, supporting applications in both humane and inhumane target categories.
    \item The proposed method is model-agnostic across its entire pipeline, offering high flexibility in model substitution and allowing easy adaptation of alternative models.
\end{itemize}
\section{Related Work}
\label{sec:related}

\begin{figure*}[t]
  \centering
   \includegraphics[width=0.9\textwidth]{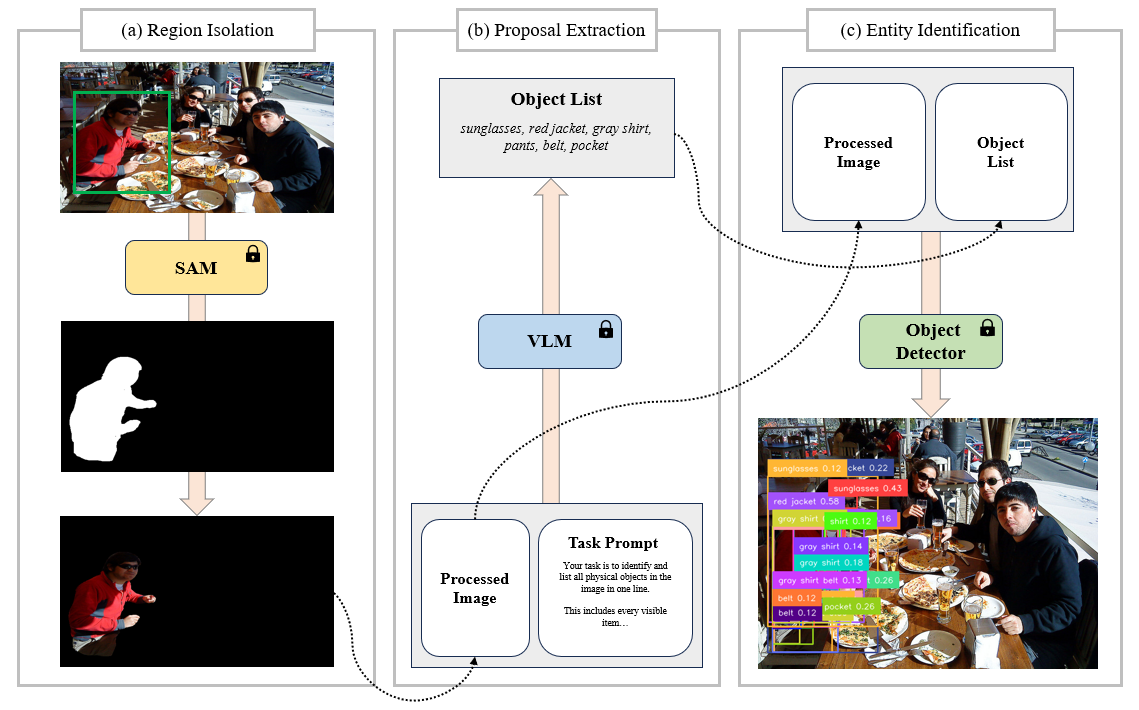}
   \caption{A \textbf{full overview} of the FOCUS framework is shown above. The first step extracts the binary mask of the region designated by the user as shown by the green box on the top left. Following this step, the VLM processes the image to output a proposal of objects. Finally, the detection model processes the proposal from the previous step to output the final detection results.}
   \label{fig:pipeline_full}
\end{figure*}

\subsection{Open-Vocabulary Object Detection}
Early studies on object detection were constrained by the limited number of objects that the models could detect and made attempts to enhance detection performance simply by scaling the size of the dataset to increase exposure to as many classes as possible \citep{zhang2023recognize, yolov3, NIPS2015_14bfa6bb}. Nevertheless, these early models struggled to generalize beyond the fixed set of categories they were initially trained on, showing weakness in their applicability to real-world scenarios due to the persisting issue of overfitting to seen categories. A key development that addressed this issue was the use of vision-language models \citep{pmlr-v139-radford21a, pmlr-v139-jia21b} to map visual and semantic spaces by utilizing datasets that consists of a large number of image-text pairs. More recent models deploy attention-based mechanisms to further enhance the link between the text queries with specific visual regions from the image \citep{liu2023grounding, ren2024grounded, Yao_2024_CVPR, Cheng2024YOLOWorld} as a default setting.

Despite the promising improvements, object detection models today still face a common paradox: whether to allow users to specify detection targets \citep{zhang2023recognize}, thereby increasing the likelihood of identifying desired objects, or to rely on zero-shot detection \citep{Hou_2024_CVPR, hou2024relation, zong2023detrs} to automate the detection process and reduce manual work. These models typically require external inputs from the users such as object lists to pinpoint the location or the label of the target. Moreover, they often output detection results that do no align with the user intention yet leave users no option to modify the process. Our work, on the other hand, builds on these advances by offering category-level adaptability independent of the complexity of the scene while providing flexibility in guiding the detection process, thereby addressing the aforementioned limitations of existing approaches.

\subsection{Vision Language Models}
Vision-language models have emerged as a cornerstone in enabling models to understand and reason about visual inputs in natural language. While these models lack the ability to locate objects without an actual detection component, their improved capacity to align textual and visual representations has greatly enhanced the performance of open-vocabulary object detection models \citep{Cheng2024YOLOWorld, Yao_2024_CVPR}. However, even with the integration of VLMs like CLIP \citep{pmlr-v139-radford21a}, existing models remain limited in processing detailed requests. Specifically, they often struggle to handle lengthy, ambiguous textual inputs that users might rely on to perform more complex or nuanced tasks, highlighting a gap in their ability to address real-world challenges effectively.

Recent works, similar to proposed method, integrate larger VLMs to its pipeline to handle more ambiguous textual prompts and more complex visual reasoning tasks not only in the image domain \citep{you2024ferret, xiao2024grounding, Lai_2024_CVPR} but also in the video \citep{yan2024visa} and robotics domain \cite{jin2024reasoning}. These approaches leverage the advanced reasoning capabilities that larger models possess to interpret nuanced prompts and generate more accurate visual insights. Yet, all these works tackle tasks that are fundamentally different from ours, open-vocabulary object detection. Furthermore, the issue of handling complex scenes persists even with these new attempts. Unlike these methods, FOCUS addresses this limitation by implementing a novel module that eliminates necessary details to enable precise detection even in complex scenes while also avoiding the need for manually curating object lists.

% \textbf{Semantic Segmentation.} Similar to open-vocabulary object detection, semantic segmentation aims to segment objects based on textual input provided by the user. With the success of VLMs like CLIP \citep{pmlr-v139-radford21a} and the Segment Anything Model \citep{kirillov2023segany, ravi2024sam2}, more approaches have been explored to lead the segmentation results to align with the user intentions. Some works take an interactive approach \citep{semantic_sam, zhao2024graco, wei2023semantic} to allow users to control and edit the segmentation process until the results align with their intentions. More recently, the field of open-vocabulary semantic segmentation has focused on generalizing segmentation tasks to an open set of categories \citep{xie2024sed, Benigmim_2024_CVPR}. These works still fall short with processing a large number of images since they require extensive manual interaction from the user. Furthermore, a lot of these models only take simple textual prompts as inputs, which restricts their abilities to process ambiguous requests.
\section{FOCUS}
\label{sec:method}

\begin{table*}[t]
    \centering
    \renewcommand{\arraystretch}{1.2}
    \resizebox{\textwidth}{!}{%
        \begin{tabular}{ccccccccc}
            \toprule
            \multirow{2}{*}{\textbf{Methods}} 
            & \multicolumn{4}{c}{\textbf{Granular}} 
            & \multicolumn{4}{c}{\textbf{Vehicles}} \\
            \cmidrule(lr){2-5} \cmidrule(lr){6-9}
            & \textbf{0.2} & \textbf{0.4} & \textbf{0.6} & \textbf{0.8} 
            & \textbf{0.2} & \textbf{0.4} & \textbf{0.6} & \textbf{0.8} \\
            \midrule
            GroundingDINO \cite{liu2023grounding} 
            & 0.465 & 0.168 & 0.042 & 0.008 
            & 0.718 & 0.427 & 0.105 & 0.018 \\
            OWLv2 \cite{minderer2024owlv2} 
            & 0.393 & 0.176 & 0.047 & 0.006 
            & 0.423 & 0.239 & 0.083 & 0.009 \\
            OpenSeeD \cite{zhang2023simple} 
            & 0.159 & 0.099 & 0.040 & 0.012 
            & 0.206 & 0.144 & 0.091 & 0.032 \\
            \midrule
            GroundingDINO + FOCUS 
            & 0.835 & 0.670 & 0.301 & 0.096 
            & 0.799 & 0.478 & 0.183 & 0.044 \\
            OWLv2 + FOCUS
            & 0.430 & 0.299 & 0.123 & 0.013 
            & 0.418 & 0.228 & 0.096 & 0.023 \\
            \bottomrule
        \end{tabular}%
    }
    \caption{\textbf{F1 Score} comparison of existing baselines against our proposed method on human images. We use GPT-4o as the VLM.}
    \label{tab:f1scores}
\end{table*}

This section introduces the FOCUS framework and details each module it undergoes to achieve its functionality. At large, FOCUS consists of three distinct stages: (i) Region Isolation, (ii) Proposal Extraction, and (iii) Entity Identification. Alongside, we aim to explain how FOCUS can take user intent into consideration while more accurately identifying fine-grained components from larger objects, especially in complex scenes that contain multiple distinct items that divert focus from the intended target. 

\subsection{Prompt-based Segmentation}
\label{sec:attentseg}
Modern vision models \citep{Cheng2024YOLOWorld, zong2023detrs} are trained to detect objects that exhibit the most prominent features within an input image. Larger objects, therefore, are more easily captured by these models due to their dominant presence. Such bias, in turn, reduces the likelihood of smaller objects nested within the larger ones to be detected, making it challenging to detect granular components. For example, if a person contains various smaller items on them, as shown in Figure \ref{fig:pipeline_full}, the model is more likely to detect the larger ``person" object rather than focusing on identifying the finer details like ``belt" within the person. This limitation becomes increasingly pronounced as input images grow more complex, containing multiple distinct elements. As such, current models are less specialized in accurately detecting objects in subtler regions. Overcoming this challenge requires assisting the model to concentrate solely on the region users intend to examine to ultimately mitigate interference from irrelevant elements in the presented image.

The FOCUS pipeline, therefore, begins with the \textbf{Region Isolation module}, which selectively masks out regions outside of the target object. This module is performed in two distinct stages. In the initial step, the original image $I_{orig}$ is directed to SAM \citep{kirillov2023segany, ravi2024sam2} along with the bounding box coordinates, $B_{input}$, set by the user to represent the target region of interest. The binary mask $M_{binary}$ from SAM is then applied to produce the masked image $I_{atten}$ that isolates and retains only the target of interest within the region initially set by the user from the original image. This step prevents the features of unrelated objects from disturbing the detection process. Not limited to simply eliminating the unwanted regions, it also sets a crucial foundation for the Proposal Extraction stage discussed in the next section, as the vision backbone of the VLM can now focus exclusively on the area that the user specified and wishes to examine exclusively to propose objects.

\subsection{Proposal Extraction}
\label{sec:objpro}
One of the persistent limitations of existing detection models is their inability to refine detection results based on user feedback. For example, from Figure \ref{fig:pipeline_full}, if a user intends to detect the jacket on the left person, but the model instead identifies that on the right, there is no mechanism to specify or adjust the detection to focus on the left. This inflexibility can lead to frustration when users are unable to direct the model toward the exact target they intend to examine.

FOCUS tackles this issue through the \textbf{Proposal Extraction module}, where we integrate VLMs to the pipeline and take in an user-controllable task prompt $P_{task}$ that is not limited in length or scope. A sample $P_{task}$ is a text prompt like "Your task is to identify and list all physical objects in the image" with additional instructions $P_{add}$, as shown in Figure \ref{fig:pipeline_full}. While VLMs possess outstanding reasoning capabilities and visual backbones, assisting the automated curation of the target object lists, their output format is not fixed. Therefore, the additional instructions $P_{add}$ is necessary to return the object list $L_{p}$ in a format that object detection models can process, while also leaving room for more detailed task instructions.

The core benefit of this integration comes from the textual reasoning capabilities of VLMs, as changes to the input prompt $P_{task}$ can allow users to guide the object detection process to align the detection results with their intentions through prompt engineering. For example, we can change the prompt in Figure \ref{fig:pipeline_full} to be "Your task is to identify and list all body parts in the image" to perform anatomic-level detection. Rather than using different models for different detection tasks, our proposed method can perform numerous detection tasks explainable in natural language with a very simple edit to the input prompt. % See Section \ref{sec:appendix_text} in the Supplement for experimental results using different prompts.

\subsection{Entity Identification}
Although VLMs are adapted with strong vision backbones that make the previous step possible, they are still limited in understanding spacial relations. In other words, the actual location of the object cannot be obtained only with VLMs. Therefore, the FOCUS pipeline adds the \textbf{Entity Identification module} to obtain the final detection coordinates.

In this final step, the object detection model takes in the processed image $I_{atten}$ from the Attentive Segmentation stage and the object proposal $L$ from the Proposal Extraction stage to output the final coordinates $B_{output}$. It is necessary that we use detection models with open-vocabulary capabilities, as closed-vocabulary models might not be able to process $L_{p}$ as the objects in the VLM proposal are not restricted by a predefined set of vocabulary. By design, we achieve open-vocabulary capabilities and the detection module is able to output a set of bounding boxes $B_{output}$ representing the coordinates of the detected items $L_d$.
\section{Experiments}

\begin{figure*}[t]
  \centering
  \includegraphics[width=0.9\textwidth]{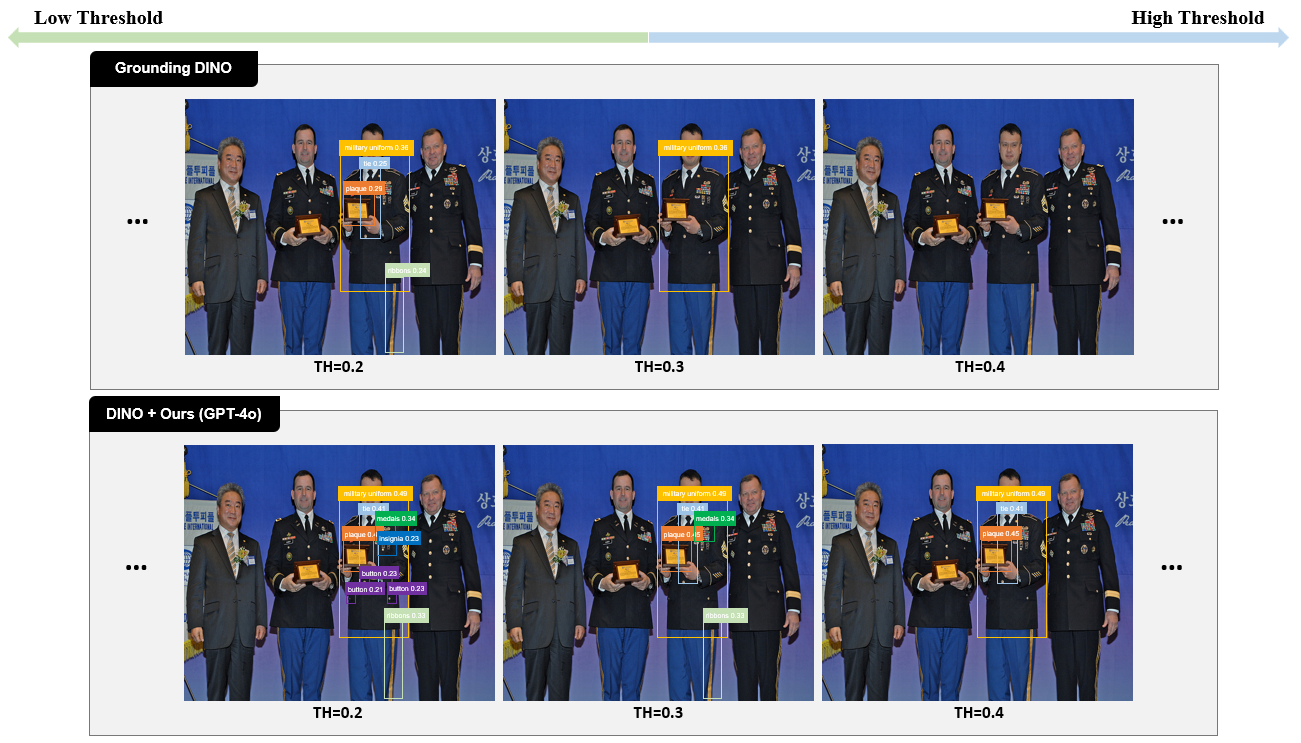}
  \caption{\textbf{Robustness} comparison between the DINO + FOCUS integration and the baseline DINO on changes to confidence threshold. Results show that FOCUS is more resistant to changes in threshold cutoffs, resulting in more confidence detections.}
  \label{fig:quantitative_3}
\end{figure*}

We now present the experimental setup and results of our FOCUS framework. We test our proposed methodology mainly on GroundingDINO \citep{liu2023grounding} and OWLv2 \citep{minderer2024owlv2} with the baseline as GPT-4o \citep{openai2023gpt4}. Evaluation of FOCUS against existing baselines on open-vocabulary object detection shows that the proposed methodology effectively enhances the detection capabilities of baseline models, both qualitatively and quantitatively.
%, with more model combinations in Section \ref{sec:ablation} in the Supplement.

\subsection{Problem Setup}
In order to test the performance of the FOCUS pipeline against existing baseline models, we need datasets that contain annotations on the targets that they wish to capture.  This testing process involves annotating hundreds of distinct entities within each image, tailored to suit each custom natural language prompt, which demands a high degree of specificity. However, the annotations available in existing datasets frequently lack the level of granularity and task-specific labels required for our specific task. Therefore, we frame our task into an quantitative alignment task between the VLM proposal and the results from the detection models. In other words, the results are evaluated based on the number of correctly detected entities while assuming the object list proposal made by the VLM to be the ground truth. To ensure fair comparison between our proposed method and existing baselines, we make two adjustments: \textbf{1)} The same object lists $L$ from the VLM output are used when testing the performance of existing methods like GroundingDINO \citep{liu2023grounding}, OWLv2 \cite{minderer2024owlv2}, and OpenSeeD \cite{zhang2023simple} and \textbf{2)} only the objects within the original bounding box $B_{input}$ are considered to be correct detections when testing baselines without the FOCUS integration. Such setup allows us to automate the testing procedure, even though the baselines are not capable of curating target lists themselves, and ensures that objects that are detected outside of the region of interest are not included during evaluation. 

Let $R$ and $P$ be the functions that calculate the recall and precision between two input lists, respectively. We calculate the F1 score as follows:
\begin{equation}
F_1(L_p, L_d) := 2 \cdot \frac{R(L_p, L_d) \cdot P(L_p, L_d)}{R(L_p, L_d) + P(L_p, L_d)}
\end{equation}

\subsection{Experimental Detail}
Our method is primarily based on the default ViT-H SAM \citep{kirillov2023segany, ravi2024sam2}, DINO \cite{liu2023grounding}, and GPT-4o \citep{openai2023gpt4}. For DINO, we rely on the official implementation and utilize its standard reference weights with the Swin Transformer \citep{liu2021Swin} visual backbone. All experiments discussed are performed on NVIDIA V100 GPUs. Although the memory requirements differ from model to model, our proposed method only requires the minimum infrastructure required to run inference on these models. The detection score threshold for the model in the Entity Identification module is set to 0.1 to output all possible detections at one execution. 

\begin{figure*}[t]
  \centering
  \includegraphics[width=0.9\textwidth]{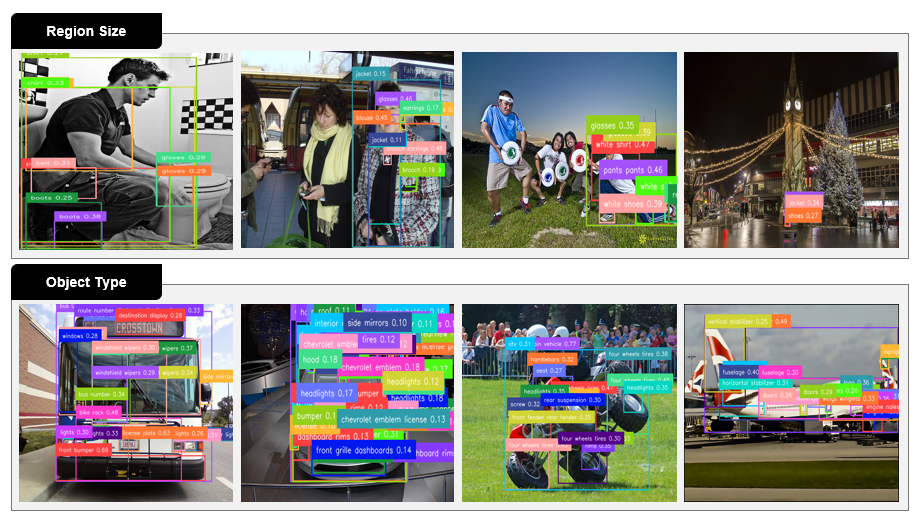}
  \caption{Sample results on different region sizes and object types. Top layer shows descending region size from left to right; bottom layer shows detections on inhumane targets. Results show that FOCUS is not restricted in the size of the target or the object type.}
  \label{fig:quantitative_1}
\end{figure*}

\begin{figure*}[!htbp]
  \centering
  \includegraphics[width=0.95\textwidth]{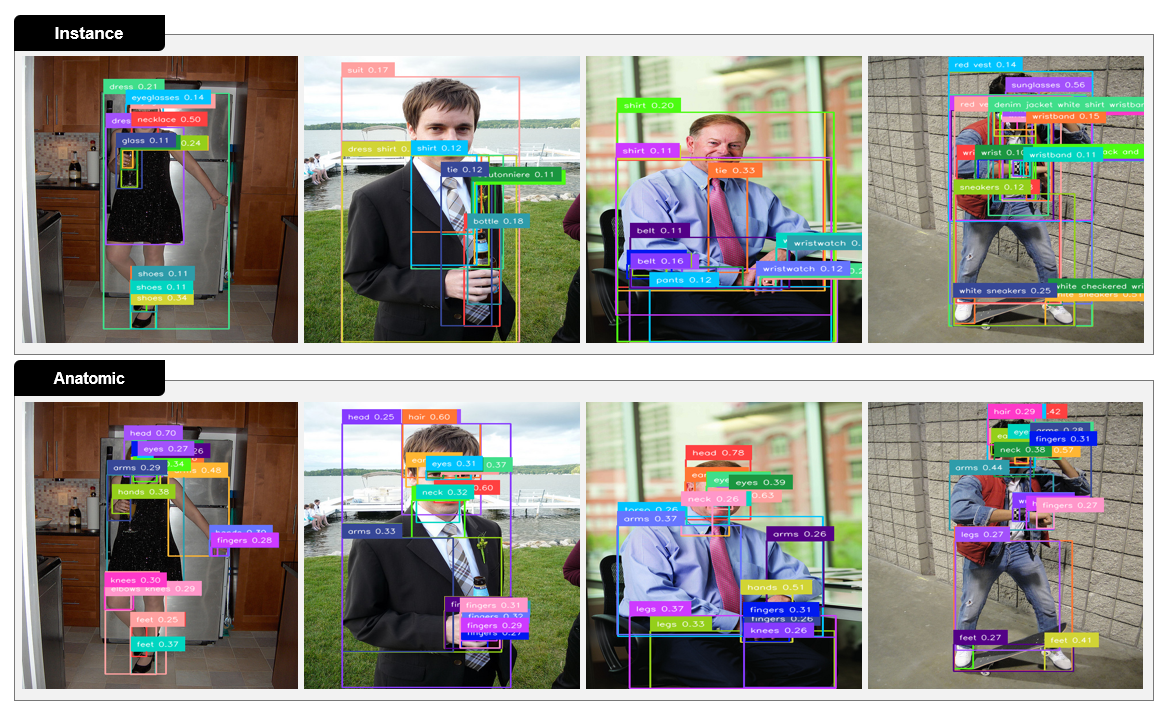}
  \caption{Sample results on the same target but using different prompts. Top later shows instance detection; bottom later shows anatomic detection. Results show that FOCUS can perform various detection tasks by varying the text prompt.}
  \label{fig:quantitative_2}
\end{figure*}

\subsection{Results}
As shown in Table \ref{tab:f1scores}, applying our FOCUS method to existing baseline models like DINO \citep{liu2023grounding} and OWL \citep{minderer2024owlv2} shows meaningful increase in performance in detecting granular components from complex scenes. For the \textbf{granular} detection task, the DINO + FOCUS combination demonstrates significant improvement of nearly 40\% in F1 score, compared to the baseline without integrating FOCUS. This increase likely comes from the Region Isolation module of FOCUS, discussed earlier in Section \ref{sec:attentseg}, that allows the baseline models to focus on the predefined region only. For the \textbf{vehicle} detection task, the difference seems relatively minimal compared to the granular detection task at first glance. We note that such results come from the nature of vehicles having more clear constituent elements than human targets. However, we can see that the performance increases with higher cutoff scores, bolstering the idea that FOCUS indeed is more resistant to changes in the cutoff thresholds, as also qualitatively shown in Figure \ref{fig:quantitative_3}.
\section{Discussions}

\subsection{Object Detection with Varying Targets}
As discussed in Section \ref{sec:intro}, one of the main contributions of our proposed method is consistent performance across various target classes. To verify this point, we test our pipeline on two main target categories. We use the 2017 COCO validation set \citep{cocodataset} for humane targets and the 2012 PASCAL VOC validation set \citep{Everingham15} for inhumane targets, mainly focusing on vehicles. Both datasets include annotations of class labels representing object categories within each image. We utilize this annotation to filter out images that contain at least one relevant label (e.g. human, vehicles). % See Appendix \ref{sec:dataset} for details on how we preprocess these datasets to align them with our experimental task.

For all model combinations we tested, we found meaningful increases in performance by applying the FOCUS pipeline to baseline models, as shown in Table \ref{tab:f1scores}. We show that forming a downstream pipeline not only grants the model the ability to process explainable requests from the user but also substantially enhances the performance of granular object detection on different targets. Note that this performance enhancement comes from the Attentive Segmentation stage of the pipeline because the experiment was controlled in a way that the only difference between the baselines and and proposed methodology was the existence of this module. In Section \ref{sec:attentseg}, we pointed out that existing baselines often face difficulties in detecting granular components in complex scenes. However, we can see that such issue is mitigated from using the proposed method.

\subsection{Object Detection at Varying Criteria}
As discussed in Section \ref{sec:objpro}, the proposed method includes VLMs to its pipeline to enable users to make explainable requests. We noted that this allows users to curate $P_{task}$ provide detailed requests to perform complex detection tasks that otherwise would not have been possible. However, the benefit of being able to provide natural language instructions goes far beyond simply providing detailed requests.

Instead of describing the task in greater detail, users can explain a different detection task to perform different types of detection even on the same target with variations to the natural language prompt. For example, referring to \ref{fig:quantitative_2}, one might make a request on instance-level detection with the intention of identifying all components that form the target person as shown in the top layer, while another user might want to detect the visible human body parts instead of the distinct physical objects as shown in the bottom layer. Our proposed method allows users to explain their requests and make such distinctions through simple edits to the prompt. In other words, FOCUS supports a wide range of instructions as long as it can be explained in natural language. % We place the quantitative results of this experiment in Section \ref{sec:promptvary} in the in the Supplement due to space limit.

\begin{figure}[t]
  \centering
   \includegraphics[width=\linewidth]{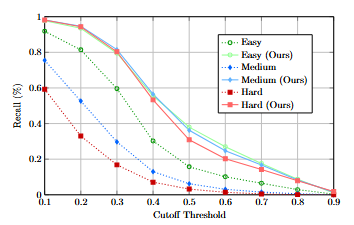}
   \caption{Here we show the performance of our model on images of varying difficulty level. We use GPT-4o as the VLM and GroundingDINO as the detector.}
   \label{fig:graph}
\end{figure}

\subsection{Object Detection at Varying Complexity}
Earlier in Section \ref{sec:attentseg}, we used the term "complex scenes" without specifying the definition other than an image containing "multiple distinct items". In this section, we formalize the definition of the phrase and share testing results on groups of images at varying levels of complexity. We first make the assumption that with the increase in the number of human targets in the image, the area of occupied by each individual target will decrease and therefore more objects will exist in the scene. This is, images with more human targets will therefore be more difficult to perform fine-grained detections due to the number of distinct elements and the decreasing size of them. As proof of concept, we list the top 10 items with highest score discrepancies that appear more than 100 times across the human target test images. % and analyze the them in Section \ref{sec:scorediff} in the Supplement.

Based on this assumption, we group the images by the number of people present in the image and test the performance of our proposed framework after grouping the dataset by levels of difficulty. Formally, let $p$ represent the number of people in an image. We categorize the dataset into three levels of difficulty as follows:
\[
\begin{cases} 
\text{Easy} & : p \leq 2 \\
\text{Medium} & : 3 \leq p \leq 7 \\
\text{Hard} & : p \geq 8 
\end{cases}
\]
We evaluate GroundingDINO \cite{liu2023grounding} against the GroundingDINO + Ours (GPT-4o) combination on all three groups, as shown in Figure \ref{fig:graph}. Results show that our proposed method outperforms the baseline by a meaningful margin and shows much more resistance to the increasing complexity of the scene. Furthermore, the performance gap widens as the difficulty increases, indicating the effectiveness of our approach in handling more challenging scenarios.
% Refer to Section \ref{sec:imgcomplex} in the Supplement for the quantitative experimental results of this experiment.

\subsection{Ablation Study}
We conduct an ablation study on the granular detection task using our default GroundingDINO + FOCUS model to evaluate the impact of the Prompt-based Segmentation module. We tested two alternative approaches to validate the use of SAM \citep{kirillov2023segany}: \textbf{1)} retaining the entire bounding box region as is and \textbf{2)} cropping the bounding box region directly from the image. Experimental results reveal that the FOCUS framework improves performance on the granular detection task by $N$\% compared to the baseline itself. While direct quantitative comparisons across these approaches are infeasible due to variations in VLM proposals, our findings indicate that retaining the entire bounding box region preserves unnecessary visual context, particularly in crowded scenes. This highlights the effectiveness of the segmentation technique in enabling models to focus more precisely on the designated region and target of interest.

% Refer to Section \ref{sec:samjustify} in the Supplement for more details on this experiment and justification on using SAM as the default method.
\section{Conclusion}
We introduce the Fine-Grained Open Vocabulary Object Detection via User-Guided Segmentation (FOCUS) framework. The proposed method enhances the capabilities of existing object detection models to extract more fine-grained components from user-designated targets, especially in complex scenes. Our work not only automates open-vocabulary object detection but also enable users to directly guide the detection process in natural language by incorporating VLMs to its pipeline. The FOCUS framework is also compatible with various object detection models. Meaningful improvements from existing baselines emphasizes the effectiveness and flexibility of our model at adapting to and performing various detection tasks. To the best of our knowledge, our work is the first to integrate VLMs in object detection to enable automated and controllable open-vocabulary object detection. We hope our work sets a foundation for future works that focus on more advanced applications.
{
    \small
    \bibliographystyle{ieeenat_fullname}
    \bibliography{main}
}


\end{document}